\documentclass{article}

\usepackage[nonatbib, preprint]{neurips_2021}

\usepackage[utf8]{inputenc} % allow utf-8 input
\usepackage[T1]{fontenc}    % use 8-bit T1 fonts
\usepackage{hyperref}       % hyperlinks
\usepackage{url}            % simple URL typesetting
\usepackage{booktabs}       % professional-quality tables
\usepackage{amsfonts}       % blackboard math symbols
\usepackage{nicefrac}       % compact symbols for 1/2, etc.
\usepackage{microtype}      % microtypography
\usepackage{xcolor}         % colors

\usepackage{biblatex}
\usepackage{graphicx}
\usepackage{float}

\title{Does MAML Only Work via Feature Re-use? A Data Centric Perspective}

\author{%
  Brando Miranda
  \\
  Department of Computer Science\\
  University of Illinois Urbana-Champaign\\
  Urbana, IL 61801 \\
  \texttt{miranda9@illinois.edu} \\
    \And
    Yu-Xiong Wang
  \\
  Department of Computer Science\\
  University of Illinois Urbana-Champaign\\
  Urbana, IL 61801 \\
  \texttt{yxw@illinois.edu}
  \And
    Sanmi Koyejo
  \\
  Department of Computer Science\\
  University of Illinois Urbana-Champaign\\
  Urbana, IL 61801 \\
  \texttt{sanmi@illinois.edu}\\
}

\begin{document}

\maketitle

\begin{abstract}
   Recent work has suggested that a good embedding is all we need to solve many few-shot learning benchmarks.
   Furthermore, other work has strongly suggested that Model Agnostic Meta-Learning (MAML) also works via this same method - by learning a good embedding. 
   These observations highlight our lack of understanding of what meta-learning algorithms are doing and when they work.
   In this work, we provide empirical results that shed some light on how meta-learned MAML representations function.
   In particular, we identify three interesting properties:
   1) In contrast to previous work, we show that it is possible to define a family of synthetic benchmarks that result in a low degree of feature re-use - suggesting that 
   current few-shot learning benchmarks {\em might not have the properties} needed for the success of meta-learning algorithms;
   2) meta-overfitting occurs when the number of classes (or concepts) are finite, and this issue disappears once the task has an {\em unbounded} number of concepts (e.g., online learning);
   3) more adaptation at meta-test time with MAML does not necessarily result in a significant representation change or even an improvement in meta-test performance - even when training on our proposed synthetic benchmarks.
   Finally, we suggest that to understand meta-learning algorithms better, we must go beyond tracking only absolute performance and, in addition, formally quantify the degree of meta-learning and track both metrics together.
   Reporting results in future work this way will help us identify the sources of meta-overfitting more accurately and help us design more flexible meta-learning algorithms that learn beyond fixed feature re-use. %(as a scientific community)
   Finally, we conjecture the core challenge of re-thinking meta-learning is in the design of few-shot learning data sets and benchmarks - rather than in the algorithms, as suggested by previous work.
   %In this work we show how reporting results in this way can help us understand meta-learning algorithms better - for example we identify met
\end{abstract}

\section{Introduction}

Few-shot learning is a research challenge that assesses a model's ability to quickly adapt to new tasks or new environments.
This has been the leading area where researchers apply meta-learning algorithms - where a strategy that learns to learn quickly is likely to be the most promising.
However, it was recently shown by Tian et al. \cite{Tian2020} that a model with a good embedding is able to match and beat many modern sophisticated meta-learning algorithms on a number of few-shot learning benchmarks.
In addition, there seems to be growing evidence that this is a real phenomena \cite{Chen2019, Chen, Dhillon2019, Huang2019}.
Furthermore, analysis of the representations learned by Model Agnostic Meta-Learning (MAML) \cite{maml} (on few-shot learning tasks) revealed that MAML mainly works by learning a feature that is re-usable for many tasks \cite{Raghu} -- what we are calling a good embedding in this paper.
%~\yx{I think we probably need to give some explanation on MAML? Also, I think we need to explain what we meant by feature re-use.}

These discoveries reveal a lack of understanding on when and why meta-learning algorithms work and are the main motivation for this work. 
In particular, our contributions are:

\begin{enumerate}
    \item We show that it is possible to define a synthetic task that results in lower degree of feature re-use, thus suggesting that 
   current few-shot learning benchmarks might not have the properties needed for the success of meta-learning algorithms;
   \item Meta-overfitting occurs when the number of classes (or concepts) are finite, and the issue disappears once the tasks have an unbounded number of concepts;
   \item More adaptation for MAML does not necessarily result in representations that change significantly or even perform better at meta-test time.
\end{enumerate}

\section{Unified Framework for Studying Meta-Learning and Absolute Performance}\label{metric_ml}

We propose that future work on meta-learning should not only report absolute performance, but also quantify and report the degree of meta-learning.
We hypothesize this is important because previous work \cite{Tian2020} has observed that supervised learning (while only fine-tuning the final layer) is sufficient to solve current meta-learning benchmarks.
This might give the potentially false impression that current trends in meta-learning are irrelevant.
To avoid that, we hypothesize that measuring the degree of meta-learning we provide in this section will provide a step forward in explaining those important observations.  
% \sk{this is too early without any definitions or explanations.}
% This is crucial not only for a better understanding of meta-learning algorithms, but also because the eventual goal is to build Artificial General Intelligence (AGI), and for that it is imperative we make deliberate efforts to measure and define it in actionable ways.~\yx{Not sure if mentioning AGI here goes too far?}
% In addition, for us to be able to understand and trust such a system, we need metrics that can diagnose basic issues,  e.g. if the system is meta-overfitting -- defined as when the system has a high degree of meta-learning \textit{coupled} with a high gap between meta-train and meta-test errors.\sk{I think this first paragraph is too early, suggest moving it later.}

%a humble but valuable first step - inspired by \cite{Raghu} - by defining
In this work, we make an important first step by emphasizing the analysis done by \cite{Raghu}, by defining the degree of meta-learning as the normalized degree of change in the representation of a neural network $nn_{\theta}$ after using meta-adaptation $A$:
\begin{equation}\label{eq_ml}
    ML(nn_{\theta}) = \mathrm{Diff}( nn_{\theta},  A(nn_{\theta}) ).
\end{equation}
In this work we set $ML(nn_{\theta})$ to be distance based Canonical Correlation Analysis (dCCA) \cite{Morcos}.
Note that dCCA is simply 1 minus CCA to switch the similarity based metric to a difference based metric between 0 and 1.

\section{Benchmarks that Require Meta-Learning}

\subsection{Background}

\textbf{Model-Agnostic Meta-Learning (MAML).}
The MAML algorithm \cite{maml} attempts to meta-learn an initialization of parameters for a neural network that is primed for quick gradient descent adaptation. 
It consists of two main optimization loops: 1) an outer loop used to prime the parameters for fast adaptation, and 2) an inner loop that does the fast adaptation.
During meta-testing, only the inner loop is used to adapt the representation learned by the outer loop.

\textbf{Feature re-use.} 
In the context of MAML, this term usually means that the inner loop provides little adaptation during meta-testing, when solving an unseen task.
In particular, Raghu et al. \cite{Raghu} showed that MAML has little representation change as measured with CCA and CKA after adaptation, during meta-testing on the MiniImageNet few-shot learning benchmark.

\subsection{Motivation for Our Work}

The analysis by Raghu et al. \cite{Raghu} showing that MAML works mainly by feature re-use is the main motivation for our work.
However, we argue that their conclusion is highly dependent on the data set (or benchmark) used.
%\sk{benchmark is a bit vague here. You mean dataset?}
This motivates us to construct a different benchmark and show that by {\em only} constructing a different benchmark, we can exhibit lower degrees of feature re-use in a statistically significant way.
%  \yx{I think this section can be made stronger by explaining more background, context, and our assumption -- First, how is the experiment in the previous work conducted, on which benchmark, what is their conclusion. And then our assumption is that their conclusion is highly dependent on the benchmark that is used. This motives us to construct a different benchmark.}
Therefore, our goal will be to show a lower degree of feature re-use than them.
% Our goal in this section is to define a benchmark that requires meta-learning (and not only a good embedding with feature re-use) to be solved effectively.
% This will be achieved by showing that a neural network meta-trained with MAML on this new benchmark has a higher degree of meta-learning (as measured by Equation \ref{eq_ml} and dCCA) than previous work.
In particular, their work \cite{Raghu} showed that the representation layer of a neural network trained with MAML had a dCCA of $0.1 \pm 0.02$ \cite{Raghu}.
{\em Therefore, our concrete goal will be to show that the dCCA on our task is greater than $0.12$}.
If this is achieved, it is good evidence that this new benchmark benefits from meta-learning and can be detected at a higher degree than previous work \cite{Raghu} in a statistically significant way.
This is our main result of this section and is discussed in detail in Section \ref{main_result}.

\subsection{Synthetic Task that Requires Meta-learning}

\subsubsection{Overview and Goal}\label{goals}
The main idea is to sample functions to be approximated, such that the final layer needs little or no adaptation, but the feature layers require a large amount of adaptation.
This type of task would forcibly require that the meta-learner learns a representation that requires the feature layers to change to achieve good meta-test performance (i.e., it cannot rely solely on feature re-use).
Therefore, to perform well, not only would it be good to adapt the representation layers, but additionally performance is likely to be obtained from a (meta-learned) initialization that is primed to change flexibly.
% In other words, tasks must not all have the same shared representation to be solved for meta-learning to be most useful and detectable.
% \sk{simplify and expand this sentence.}
In summary, our goal will be to construct synthetic tasks such achieving high meta-test performance and detectable meta-learning - as discussed in section \ref{metric_ml} - one needs to go beyond feature sharing.

\subsubsection{Definition}\label{def}

In this section, we describe a family of benchmarks that exhibits detectable meta-learning and requires more than a re-usable representation layer to be solved.
We propose a set of regression functions specified as a fully connected neural network (FCNN), such that the magnitude of parameters of the representation are larger than the head.
In particular, we sample the parameters of the representation layer from a Gaussian with a larger standard deviation, compared to the parameter sampling of the head.
We define the representation layer to be the first $L-1$ layers, and the head to be the final layer.
% We conjecture this encourages our meta-learner to learn a representation that benefits from adaptation to solve the task. 

Next, we describe the process to sample one function (regression task
%\sk{regresion task description should come earlier}
) from a Gaussian distribution.
We have two pairs of benchmark parameters $[(\mu^{(1)}, \sigma^{(1)}), (\mu^{(2)}, \sigma^{(2)})]$:
$(\mu^{(1)}, \sigma^{(1)})$ to sample the parameters for the representation layer, 
and $(\mu^{(2)}, \sigma^{(2)})$ to sample the parameters for the final layer.
Then each regression task $f^{(t)}$ (with index $t$) is sampled as follows: %~\yx{need to explain t.}:
\begin{itemize}
    \item Sample the representation parameters $w^{(l)} \sim N(\mu^{(1)}, \sigma^{(1)})$ for each layer $l \in [L-1]$ in the representation layers
    \item Sample the final layer parameters $w^{(L)} \sim N(\mu^{(2)}, \sigma^{(2)})$
\end{itemize}

\begin{figure}[ht]
\centering
\includegraphics[width=0.46\linewidth]{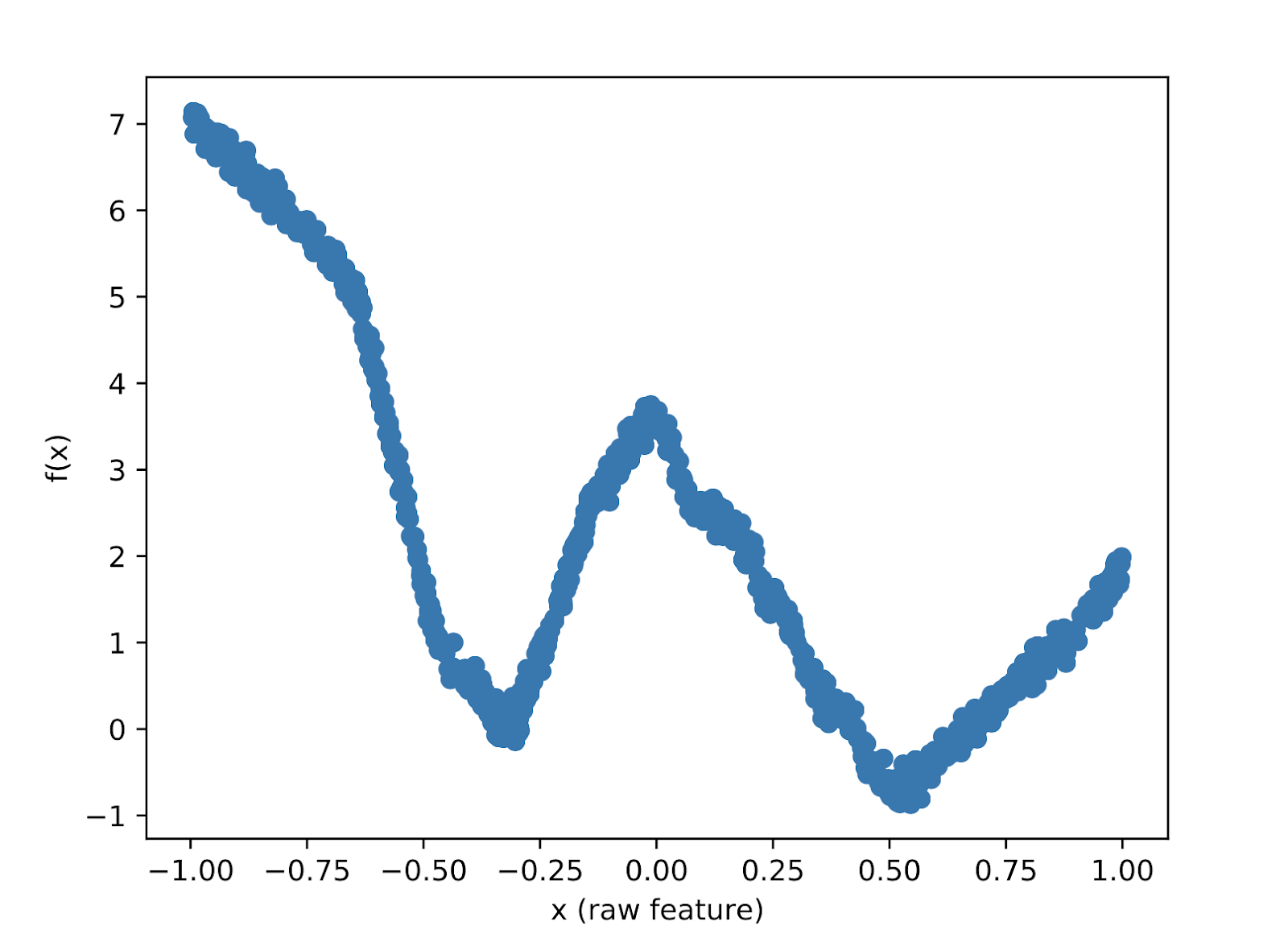}
\caption{An example regression task constructed as described in Section \ref{def}. Addressing such tasks requires high degree of meta-learning.}
\label{fun_reg}
\end{figure}

The idea is that for some $c \in \mathbb R$ we have $\sigma^{(1)} > c \cdot  \sigma^{(2)}$ such that the variance in tasks is due to the representation layers, and therefore adapting the representation layers is necessary.
For all our experiments $\sigma^{(2)}=1.0$.
An example task can be seen in Figure \ref{fun_reg}.
During meta-training, points are uniformly sampled from $[-1,1]$, and the standard support set and query set are constructed by computing $f^{(t)}_{w}(x)$.

\subsubsection{Results on Benchmarks that Require Meta-Learning}\label{main_result}

\begin{figure}[ht]
\centering
\includegraphics[width=0.63\linewidth]{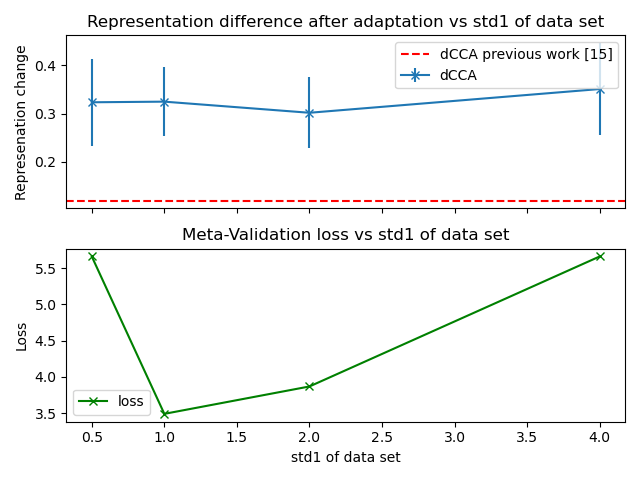}
\caption{
Shows the of lack of feature re-use and a higher degree of meta-learning, as the standard deviation of the representation layer $\sigma^{(1)}$  for generating regression.
The x-axis is the standard deviation (std) of the parameter $\sigma^{(1)}$ for generating the tasks for the data sets. 
The models used for each point in the plot are models selected from early stopping (using the meta-validation MSE loss) when meta-trained with MAML.
The models are the same architecture as the target function (4 layers fully connected neural network) with ReLU activation function. 
We also show the meta-validation loss vs the standard deviation of the task.
The dCCA was computed by from the average and standard deviation over the representation layers, in this case the first three layers.
The average is across different runs using the same meta-learned initialization.
% The red dotted line shows the the mean plus the standard deviation of previous work.
The red dotted line shows the value of $0.12$ that our models have to be statistically significant.
The only difference of this figure with respect to figure \ref{relu_metaoverfitted} is that we selected a model with the best validation here and in the figure \ref{relu_metaoverfitted} we selected the model in last step.
}
\label{best_relu_vs_std}
\end{figure}

In this section, we show a higher degree of meta-learning and a lower degree of feature re-use from an initialization trained with MAML on the benchmarks described in Section \ref{def}. 
In particular, we show this in Figure \ref{best_relu_vs_std} because the dCCA value exhibited is much larger than $0.12$ of previous work \cite{Raghu}.
{\em Most importantly, the results are statistically significant, because the error bars do not intersect with the red dotted line with (worst case) dCCA value of $0.12$. }
The red dotted line is the top error band of previous work - i.e. the mean plus the standard deviation.

\begin{figure}[ht]
\centering
\includegraphics[width=0.63\linewidth]
                   {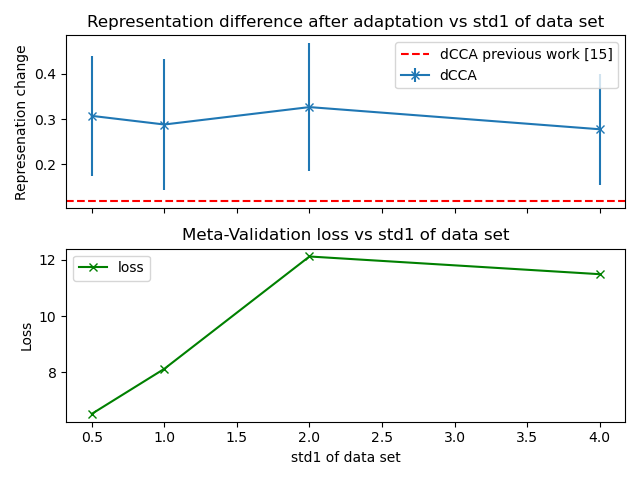}
\caption{
This figure supports the main result of the paper because a higher degree of meta-learning and a lack of feature re-use are present -- even in models that are meta-overfitted.
A meta-overfitted model can be easily obtained in our experiments by selecting a model at the final iteration.
The x-axis is the standard deviation (std) of the parameter $\sigma^{(1)}$ for generating the tasks for the data sets.
The red dotted line shows the value of $0.12$ that our models have to be above for statistically significant results that support our claims.
The only difference of this figure with respect to figure \ref{best_relu_vs_std} is that we selected a model in last step (after trough, and it had meta-overfitted) while in figure \ref{best_relu_vs_std} we select the model with lowest meta-validation loss.
}
\label{relu_metaoverfitted}
\end{figure}

Note that a dCCA higher than $0.12$ was observed across all of our experiments in over sixteen different benchmarks. 
In particular, this happened even in models that had meta-overfitted, e.g., see Figure \ref{relu_metaoverfitted}.
This is strongly suggestive that the benchmarks we defined in Section \ref{def} require meta-learning, since they do not solely rely on feature re-use to be solved. %~\yx{The setup difference between Fig2 and Fig3 is not clear.}

\section{Meta-Overfitting} \label{meta_overfitting}

In this section, we show how being armed with the additional metric discussed in Section \ref{metric_ml}, we are able to identify an increasing gap between the meta-test and meta-train losses/accuracy -- a term we refer to as \textit{meta-overfitting}.
In particular, this phenomenon is observed when we meta-train models with MAML, and becomes more pronounced as the number of iterations increases.
We attribute this to the adaptation, because this increase in the meta-generalization gap is observed in conjunction with the low degree of feature re-use (as discussed in Section \ref{main_result}), and is most noticeable in our synthetic benchmarks compared to MiniImagent \cite{Raghu}. 
Note that the dCCA of the models was much larger in our synthetic benchmarks than in MiniImagent.
In addition, we show that if the number of regression tasks (in this case functions) is not fixed, then the meta-overfitting issue is no longer observed.%In particular w We believe that because 

\subsection{Finite Number of Tasks}\label{finite_metaoverfit}

\begin{figure}[ht]
\centering
\includegraphics[width=0.55\linewidth]{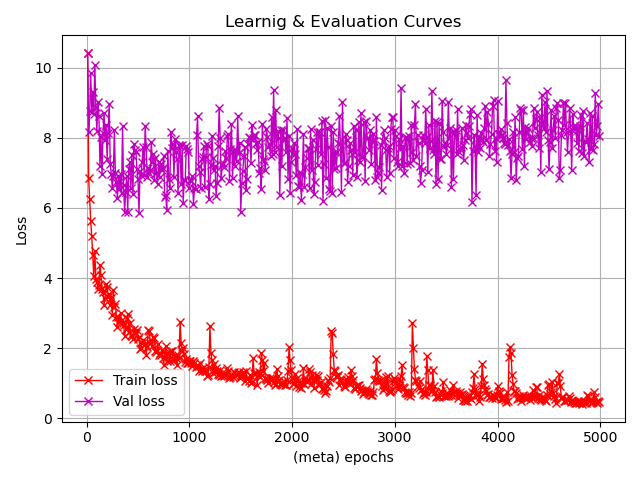}
\caption{
Shows meta-overfitting when the number of tasks (functions) is finite at $200$ regression tasks because the meta-validation loss increases as the meta-train loss decreases.
In particular, the dCCA for these models was $0.36 \pm 0.12$ corresponding to $\sigma^{(1)}=1.0$.  % from figure \ref{best_relu_vs_std}.
The plot is the learning curve for a 4-layered fully connected neural network trained with MAML \cite{maml} using episodic meta-learning.
Note that we use a (large) meta-batch size of $75$ to decrease the noise during training in the figure.
The main difference of this figure with figure \ref{no_overfit} is that in this one has a finite set of tasks using our synthetic benchmark while the other has an infinite set of tasks using the sinusoidal tasks suggested in \cite{maml}.
}
\label{meta_overfit1}
\end{figure}

When the number of regression tasks (functions) is finite ($200$ in our experiments), we consistently observe meta-overfitting.
We show this in Figure \ref{meta_overfit1} by increasing the meta-generalization gap (i.e. an increase in the difference between the meta-train and the meta-validation losses).
This is consistently observed in over $30$ experiments with a finite number of regression tasks.

\begin{figure}[ht]
\centering
\includegraphics[width=0.67\linewidth]{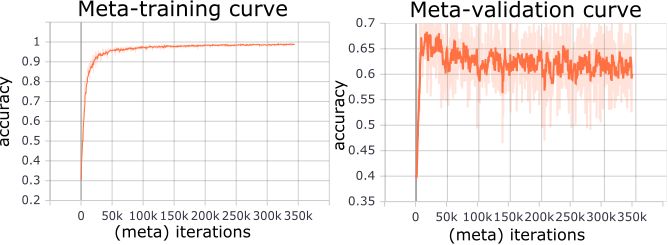}
\caption{ 
Shows that meta-overfitting is a real phenomenon in MiniImagent.
We interpret this due to the peak in the meta-validation accuracy followed by a decline as the number of iterations increases.
Importantly, the meta-train accuracy continues to increase as it converges.
The model trained is an out-of-the-box PyTorch ResNet-18.
Note that the higher noise of the meta-validation accuracy is due to having a meta-batch size of $2$ to speed up experiments.
We smoothed the meta-validation curve with a TensorBoard smoothing weight of $0.8$.
We consistently saw that increases in meta-batch size lead to decreases in noise in the learning curves, but we didn't re-run these experiments since it can take up to a week to reproduce an episodic meta-learning run - even on a Quadro RTX 6000.
% using the PyTorch libraries Torchmeta and Higher \cite{torchmeta, higher}.
}
\label{overfit_mini}
\end{figure}

Furthermore, meta-overfitting is also observed in a few-shot image recognition benchmark.
This is shown in Figure \ref{overfit_mini} with MiniImagent.
With a PyTorch ResNet-18 model, one can observe a meta-generalization gap of about $30\%$.
With a state-of-the-art ResNet-12 \cite{Tian2020}, the meta-generalization gap is instead about $20\%$.

\subsection{Infinite Number of Tasks}

\begin{figure}[ht]
\centering
\includegraphics[width=0.55\linewidth]{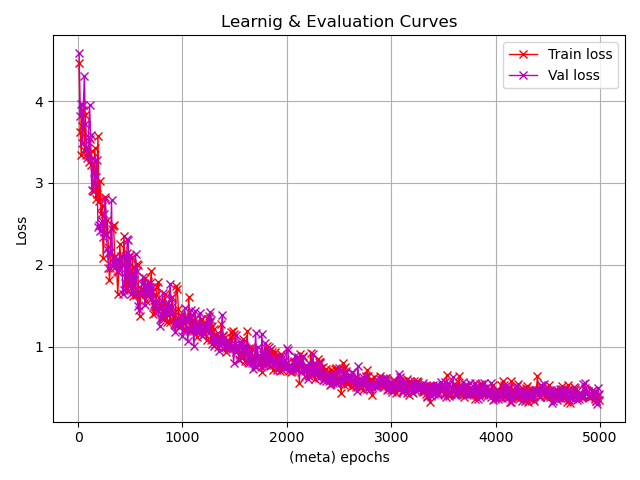}
\caption{
Shows that meta-overfitting does not occur and perfect meta-generalization occurs when the number of tasks (functions) is unbounded when training with MAML.
In other words, the meta-train and meta-validation error are indistinguishable and decrease together as the meta-iterations increases.
The main difference of this figure with figure \ref{meta_overfit1} is that in this one has a finite set of tasks using our synthetic benchmark while the other has an infinite set of tasks using the sinusoidal tasks suggested in \cite{maml}.
}
\label{no_overfit}
\end{figure}

It is interesting to highlight that meta-overfitting was not observed when the number of regression tasks is unbounded, as shown in Figure \ref{no_overfit}.
%~\yx{Here you changed the task types and benchmarks. I think we need to explain why.}
This evidence suggests that, when the number of tasks is unbounded but sampled from a related set of tasks, meta-learning algorithms can leverage their power to adapt without meta-overfitting.

To measure the amount of meta-learning and the lack of feature re-use, we compute the dCCA value of the model as in Section \ref{main_result} and observe a value of  $0.44 \pm 0.11$.
This also implies that the degree of meta-learning is higher when the number of tasks is unbounded. 

The main contribution is that this evidence suggests {\it we need to rethink how we define the few-shot learning benchmarks for meta-learning}. 
We hypothesize this is true because changing the property - like the number of concepts available to the learning - changes the behaviors of meta-learning algorithms. 
In particular, MAML stops meta-overfitting.
This suggests to practitioners that MAML is a good algorithm for online or lifelong learning benchmarks - rather than deploying it to benchmarks with a fixed number of concepts.
Overall, our evidence suggests that, as a research community, we are applying meta-learning algorithms to the wrong data sets and benchmarks.

\section{Effects of More Meta-Adaptation}

In this section, we show that increasing the number of inner steps for MAML during adaptation does not necessarily change the representation further as measured with dCCA (as in Equation \ref{eq_ml}).
In addition, the meta-validation performance also does not change.

\begin{figure}[ht]
\centering
\includegraphics[width=0.63\linewidth]{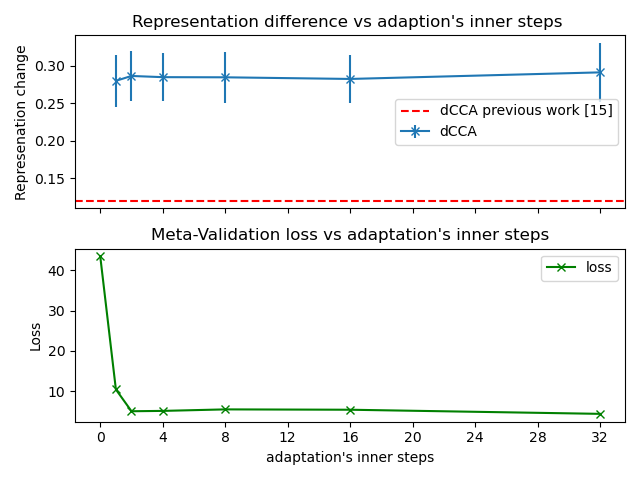}
\caption{
Shows 1) the lack of representation change and b) meta-validation change as the number of inner steps increases.
1 is shown by the relative flatness of the blue and orange lines in the upper plot.
Similarly, 2 is shown by the flatness of the green line in the lower plot.
In particular, notice that we exponentially increase the inner steps from 1 to 2 to 32.
The models used are 4 layered FCNN trained with MAML with 1 inner step and 0.1 inner learning rate, selected using early stopping using the meta-validation set with the Sigmoid activation function.
The only difference of this figure with figure \ref{ml_loss_vs_inner_steps_relu_best} is that this figure uses a sigmoid activation and the other one uses a ReLU.
Note that this is the model used for figure \ref{meta_overfit1}.
Note the dCCA value remains above 0.12 suggesting lower degree of feature re-use.
}
\label{ml_loss_vs_inner_steps_sigmoid_best}
\end{figure}

\begin{figure}[ht]
\centering
\includegraphics[width=0.63\linewidth]{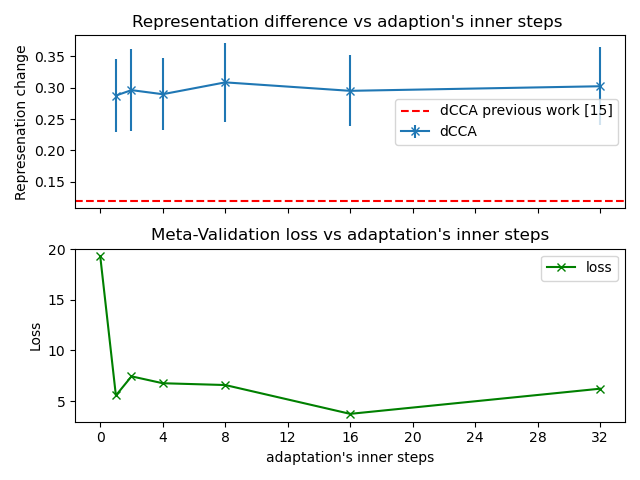}
\caption{
Shows 1) the lack of representation change and b) meta-validation change as the number of inner steps increases.
1 is shown by the relative flatness of the blue and orange lines in the upper plot.
Similarly, 2 is shown by the flatness of the green line in the lower figure.
We want to emphasize that we exponentially increase the inner steps from 1 to 2 to 32.
The models used are 4 layered FCNN trained with MAML with 1 inner step and 0.1 inner learning rate, selected using early stopping using the meta-validation set with the ReLU activation function.
The only difference of this figure with figure \ref{ml_loss_vs_inner_steps_sigmoid_best} is that this figure uses a ReLU activation and the other one uses a sigmoid.
Note the dCCA value remains above 0.12 suggesting lower degree of feature re-use.
}
\label{ml_loss_vs_inner_steps_relu_best}
\end{figure}

To show this, we obtain a single neural network meta-trained with MAML using a dataset as described in Section \ref{def}.
Then we plot how the representation changes and how the meta-validation error changes as a function of the inner steps.
We show this in Figures \ref{ml_loss_vs_inner_steps_sigmoid_best} and \ref{ml_loss_vs_inner_steps_relu_best}.
%We observe that the MAML neural networks are robust to meta-overfitting because the meta-validation loss does not increase as the number of inner steps.
We observe that the MAML neural networks are robust to meta-overfitting with respect to the inner steps of its inner adaptation rule.

Note that this is different from what was observed in Section \ref{finite_metaoverfit}, because that section shows it as a function of the meta iterations (what is sometimes called outer iterations).
In addition, it is important to emphasize that the representation change in the plots is above the $0.12$ compared to previous work \cite{Raghu}, supporting the main results of section \ref{main_result}.

\section{Related Work}\label{related_section}

% \sk{a bit awkward placement, usually before experiments or after intro. May be ok...}
Oh et al. \cite{boil} show that one can encourage models to use less feature re-use purely algorithmically by setting the inner learning rate to zero for the final layer. 
They showed BOIL outperforms MAML in both traditional few-shot learning (e.g. meta-trained on MiniImagent then meta-tested on MiniImagent) and cross-domain few-shot learning (meta-trained on MiniImagent then meta-tested on tiered ImageNet). 
In particular, their cross-domain few-shot learning is similar in spirit to the synthetic task we propose in section \ref{def}.
However, note that we show that even MAML - an algorithm that has been shown to work by feature re-use \cite{Raghu, boil} - can exhibit large representation changes if it is trained solely on a task that requires large feature changes.
Concisely, we encourage rapid learning by only changing the task, while Oh et al. \cite{boil} encourage it by changing the algorithm itself.

Guo et al.'s \cite{bscd_fsl} work is similar to ours in that they focus on defining a benchmark more appropriate for meta-learning and transfer learning. 
They propose that meta-learning should be done in a fashion where the distribution of tasks sampled changes considerably when moving from meta-training to meta-evaluation.
Our work is different in that we emphasize that the meta-training tasks themselves need to have diversity to be able to encourage meta-learning.
Although Guo et al.'s \cite{bscd_fsl} meta-evaluation procedure is excellent, we hypothesize - based on our results - that their benchmark won't have enough diversity to encourage large representation changes during meta-training.
However, we conjecture that combining our ideas and theirs is a promising step for creating a better benchmark for meta-learning.

Similar work by Triantafillou et al. \cite{Triantafillou2019} attempt to improve benchmarks by merging more data sets, but we hypothesize their data sets are not diverse enough to achieve this.
In terms of methods, our work is most similar to Raghu et al. \cite{Raghu}, but they lack an analysis of the role of the tasks in explaining their observations.
There is also other work \cite{Chen2019, Chen, Dhillon2019, Huang2019} that shows that a good representation is sufficient to achieve a high meta-accuracy on modern few-shot learning tasks e.g. MiniImagent, tiered-Imagenet, Cifar FS, FC100, Omniglot, \cite{Tian2020}, which we hope to analyze in the future.
We conjecture in is imperative that a definition of meta-learning is developed and explicitly connected to the general intelligence.
Chollet \cite{Chollet2019} takes this direction, but to our understanding the proposed definition is mostly focused on program synthesis. 
% but extension with our proposed metrics to apply for few-shot learning would be fantastic.
We also hope that in the future a metric for AI safety is ubiquitously reported as suggested in Miranda et al. \cite{foundationsmetalearning}.
% \sk{for future work, all the "believe's" in the previous related work will significantly reduce reviewer confidence in your assertions. I suggest conceptual, theoretical or empirical evidence to back it up instead.}

\section{Discussion}

% \sk{Same here, avoid "believe" unless there is a really good reason, even then use sparingly. Instead consider "hypothesize", "conjecture", or "evidence suggests", or other more scintific language}
It is exciting evidence that by only changing the few shot learning benchmark, one can consistently show higher degrees of representation changes as measured by two different metrics. 
We hypothesize this is the case because the meta-learning system has to be trained explicitly with a task that demands it to learn to adapt.

An important  discussion point is the lack of an authoritative definition for measuring meta-learning in our work and in the general literature.
In particular, in our work, we decided to not report any results with CKA.
We decided this because Ding et al. \cite{Ding} showed that it's possible to remove up to 97\% of the principal components of the weights of a layer until CKA starts to detect it.
Thus, we used dCCA which doesn't have the problem.
It instead has a higher variance, but it's easier to address this with experiment repetition sand error bars (which we did).
However, we hypothesize it would be interesting to use and extend Orthogonal Procrustes as suggest by \cite{Ding} in future work.

The most obvious gap in our work is a thorough analysis with a real world vision data set.
We hope to repeat our work with the hinted extension in section \ref{related_section} benchmarks as suggested in \cite{bscd_fsl, Triantafillou2019}. 

In addition, Figures \ref{ml_loss_vs_inner_steps_sigmoid_best}, \ref{ml_loss_vs_inner_steps_relu_best} shows that as the number of inner steps increases, the dCCA does not increase.
This is somewhat surprising given the meta-overfitting results observed in section \ref{meta_overfitting} and further experiments would be valuable.

%%%%%%%%%%%%%%%

\begin{ack}
We'd like to thank Intel for providing our team with access to their Academic Cluster Environment (ACE). 
Their computational resources and support from their staff were essential to the successful completion of our project.
In addition, this work utilized resources supported by the National Science Foundation’s Major Research Instrumentation program, grant 1725729, as well as the University of Illinois at Urbana-Champaign \cite{Kindratenko2020}.
We'd like to acknowledge the work and authors of Anatome, TorchMeta and higher \cite{anatome, torchmeta, higher} for making their code available and answering ours questions in their project's GitHub repository.
We'd like to acknowledge the weights and biases (wandb) framework for powerful tracking of experiments \cite{wandb}.
We acknowledge the feedback from Open Philanthropy on the proposal on the foundations of meta-learning \cite{foundationsmetalearning} that inspired this work.
We acknowledge the anonymous reviewers from NeurIPS for the valuable feedback for this work.
\end{ack}

%%%%%%%%%%%%%%

%\section*{References}

% References follow the acknowledgments. Use unnumbered first-level heading for
% the references. Any choice of citation style is acceptable as long as you are
% consistent. It is permissible to reduce the font size to \verb+small+ (9 point)
% when listing the references.
% Note that the Reference section does not count towards the page limit.
\medskip

% {\small
% \printbibliography
% }

\printbibliography

\appendix

\section{Supplementary Material}

\subsection{Code}

We heavily relied on our open source machine learning library Ultimate-Utils \cite{brando2021ultimateutils}. 
For representational analysis, we forked and heavily improved the Anatome library \cite{anatome} and named our open source version Ultimate-Anatome \cite{miranda2021ultimate_anatome}.

\subsection{Experimental and hyperparameter details}

\subsubsection{Details for experiments on our benchmark that requires more than feature re-use}

All models were trained on a CPU cluster with intel CPUs.
All models were 4 layered FCNN.
All models had batch-normalization and collected running statistics during meta-training, but used batch statistics during training.
MAML models for figure \ref{best_relu_vs_std} and \ref{relu_metaoverfitted} had inner learning rate of $0.1$ and $1$ inner step.
One inner super was chosen to further emphasizes the feature re-use, since it is the lowest inner step we can choose (nothing lower exists except 0 which doesn't exhibit adaptation).
Adam outer optimizer was used with learning rate $0.001$ and default parameters.
No learning schedulers were used but would be interesting to experiment with.
Since there were $200$ regression tasks, we trained the models with meta-epochs instead of meta-iterations.
This means that we reported errors, losses etc. after all tasks were seen.
Note that the input and target values were guaranteed to be novel because during meta-train and meta-testing we sample a function and generate data on the fly - similar to online learning.
Note that this is very similar to how classification tasks for few-shot work (e.g. MiniImagenet) because those tasks have a very limited number of image classes and thus results in highly correlated tasks.
In addition, we showed how both exhibited similar meta-overfitting.
For CCA value computation, we used a query set size of $100$ due to numerical issues with the Anatome library \cite{anatome}.
We did not use first order MAML.
All models were trained until convergence with about $200,000$ meta-epochs.

For models with \ref{ml_loss_vs_inner_steps_relu_best} and \ref{ml_loss_vs_inner_steps_sigmoid_best}, we meta-trained with MAML but used 2 inner steps.
The remaining hyperparameters remain the same.
We did not use first order MAML.

\subsubsection{Details for experiments on ResNet-18 meta-overfitting on MiniImagenet}

The ResNet-18 is a standard out of the box ResNet-18 downloaded from PyTorch.
We trained the models for $5,000,000$ meta-iterations.
We used $1$ inner step with $0.1$ inner learning rate.
We used meta-batch size of $4$ and $2$ for meta-training and meta-testing respectively.
We used an outer learning rate of $0.001$ and Adam with default parameters.
We did not use first order MAML.

\subsection{Role of Backbone on meta-accuracy}

\begin{figure}[ht]
\centering
\includegraphics[width=0.5\linewidth]
                   {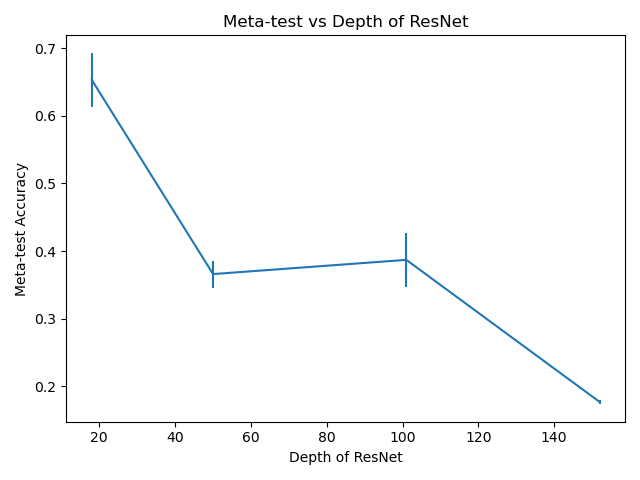}
\caption{ Shows that as the backbone of the PyTorch ResNets increases to 152 the meta-accuracy on MiniImagent decreases.
These models were trained with supervised union training is in \cite{Tian2020}.
The meta-adaption algorithm used logistic regression and was adapted to convergence on the final layer as in \cite{Tian2020}.
When using the PyTorch ResNet models instead of the special ResNets designed for MiniImagenet \cite{Tian2020} we observe see that the meta-accuracy decreases \ref{effect_of_bb_resnet18_pytorch}.}
\centering
\label{effect_of_bb_resnet18_pytorch}
\end{figure}

In this section, we describe the relation of the depth of a PyTorch ResNet model with the meta-test accuracy.
The motivation for these experiments is that if we can close the gap on MiniImagenet to over $90\%$ by only increasing the backbone depth, then this would provide strong evidence that such benchmarks really only need a good embedding.
However, we discovered that for the ResNets used in \cite{Tian2020} it seems that accuracy saturates at $80\%$ (results not shown in paper) but when using the PyTorch models we see meta-overfitting and decreasing meta-test error \ref{effect_of_bb_resnet18_pytorch}.
This suggests that even this simple scenario of few-shot learning still has space for meta-learning to be a solution.

\subsection{Analysis of meta-learned initialization}

\begin{figure}[ht]
\centering
\includegraphics[width=0.55\linewidth]{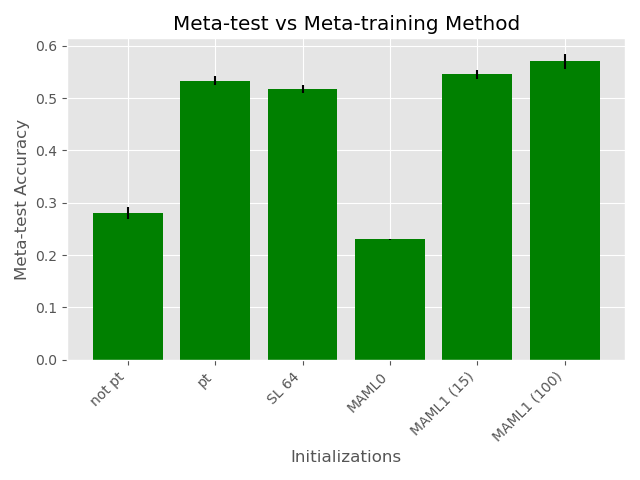}
\caption{ Shows relation of meta-test accuracy with models with a different meta-learned initialization.
PT stands for Pre-trained on Imagenet. 
Not Pt stands for a random model.
All models are ResNet18s from PyTorch. 
SL 64 stands for supervised union pre-training on MiniImagenet using all 64 labels during meta-training.
MAML0 stands for only using episodic meta-training (i.e. MAML with zero inner steps).
MAML1 (15) and MAML1 (100) stand for training using MAML with a query set of size 15 to 100. 
The meta-adaptation is the same as in \cite{Tian2020} (training logistic regression in the final layer to convergence).
}
\centering
\label{effect_of_init_resnet18_pytorch}
\end{figure}

In this section, compare meta-test accuracy of different meta-learned initialization with a PyTorch ResNet18.
We use the logistic regression adaptation used in \cite{Tian2020} at meta-test time.
The results in figure \ref{effect_of_init_resnet18_pytorch} are mixed, but it is interesting to note MAML with no inner steps performs worse than a random neural network.
This result is interesting because this is very similar to supervised pre-raining in that no meta-learner is present during training but instead of seeing all 64 images it sees 5 randomly (but uses no meta-learner).
We would have expected that the initialization obtained would have been equivalent to one with supervised pre-training.
Since they are not, it shows a MAML is at the very least capable of learning a representation that is invariant to concept permutation.

\subsection{Training with zero number of inner steps}

\begin{figure}[ht]
\centering
\includegraphics[width=0.9\linewidth]{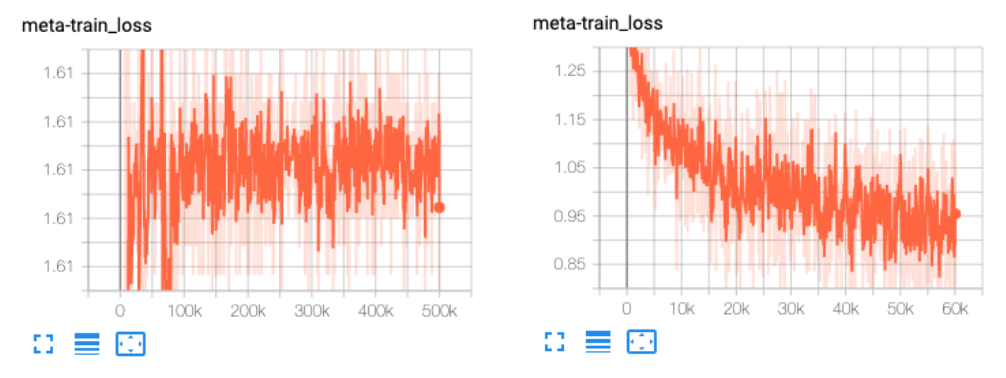}
\caption{ Compares MAML0 (only episodic training) vs MAML1 (MAML with 1 inner step).
MAML0 remains close to chance with a high loss, while MAML1 converges.
This suggests MAML0 is not equivalent to supervised pre-training and that MAML1 does learn a representation that is invariant to class order permutation.
}
\centering
\label{maml0}
\end{figure}

\begin{figure}[ht]
\centering
\includegraphics[width=0.6\linewidth]{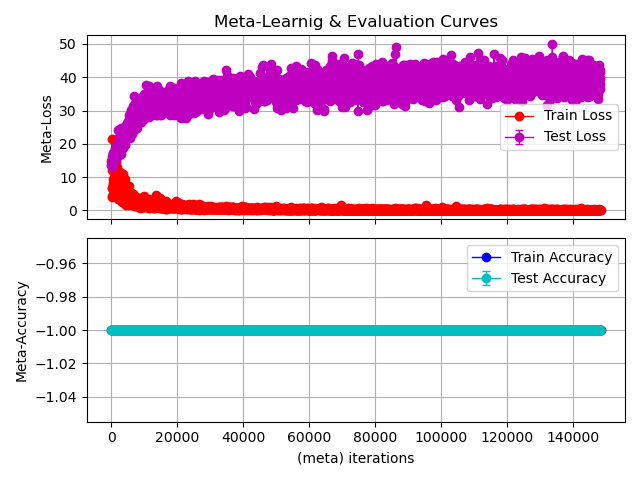}
\caption{ Shows MAML0 (only episodic training) getting zero meta-train loss (red curve) for a synthetic regression task.
This suggests that meta-learning in regression and classification might not be enitrely equivalent.
Note that meta-overfitting is still observed (purple curve).
This is a regression task, so the blue curves can be ignored.
}
\centering
\label{maml0_reg}
\end{figure}

An interesting observation is that MAML with 0 inner steps (MAML0) (i.e. only using episodic meta-training) resulted in very different meta-learned initialization compared to MAML with 1 inner step (MAML1) on MiniImagenet.
Previous work observed that supervised pre-training \cite{Tian2020} with all $64$ images during meta-training results in a strong baseline.
With this in mind, it is natural to ask: what is the difference between seeing all $64$ images during supervised pre-training or seeing only $5$ using episodic training?
With this in mind we trained MAML0 and obtained a model that performs at chance. 
Figure \ref{maml0} compares MAML0 with MAML1 to show that MAML0 obtains a model that has a very high meta-training loss.
Additionally, figure \ref{effect_of_init_resnet18_pytorch} shows such an initialization performed even worse than random.
This is surprising, but it seems that meta-learned initialization with MAML1 learn at least a model that is invariant to permutation of the order of the classes.
Unfortunately, this result seems to only be reproducible in classification, since training MAML0 in a synthetic regression task did converge to have model with low meta-train loss \ref{maml0_reg}.
This suggests future studies would be interesting to disentangle the casual factors.

\subsection{Tips and tricks for episodic meta-training}

From our experiments, we suggest the following (when episodic meta-training \cite{maml}):

\begin{enumerate}
    \item Use many query examples, e.g. greater than popular $15$ (since they often speed up convergence of the meta-learning algorithm).
    \item A large meta-batch size (since it's important to be able to have a low level of noise when tracking the meta-validation error/loss for doing early stopping).
    We found empirically for $75-100$ tasks to be a good meta-batch size.
    \item Episodic training as suggested in \cite{maml} is expensive and takes at least a week to train on MiniImagenet on a Quadro RTX 6000 using TorchMeta and higher \cite{torchmeta, higher}, so these suggestions are important.
    \item Experiments with synthetic data were run with CPU only.
\end{enumerate}

\subsection{Future work}

% \subsection{Confirming meta-learning is happening}

  \subsubsection{Summary}

\begin{enumerate}
    % \item We need to compare the amount of rapid learning (measured via CCA) more carefully in the case where $\sigma^{(1)} < \sigma^{(2)}$ bellow the current $0.5$
    % (since this case is where a fixed embedding is enough to solve a task sampled form our synthetic benchmarks).
    % \item We need to compare the following inequality: $ CCA( A(f_{maml}), f_{maml} ) > CCA( A(f_{sl}), f_{sl})$. 
    % If the inequality holds then it's true that the rapid learning of $f_{maml}$ is larger than that of $f_{sl}$, since showed a larger representation change.
    \item Defining a synthetic benchmark that is a classification problem that also requires meta-learning (or rapid learning with MAML).
    \item We also hope to construct a (real) benchmark from images that requires meta-learning.
    Formally, we propose a good start would be a benchmark where the probability of two tasks having the same class be small, otherwise we are more likely to see overfitting.
    Alternatively, a benchmark that requires the tasks to be different by at least requiring a different representation.
    We hypothesize compositionality is an ideal benchmark since this would allow sophisticated re-use of lower level representations and 
    simultaneously have an unbounded number of tasks.
    Humans are able to richly and flexibly cope with both.
    Additionally, it would be interesting to be able to quantify the distance between two different N-way, K-shot tasks to make these ideas more rigorous.
    \item An interesting benchmark with many classes with real images is taking the union of many vision classification tasks and re-scaling all images to be of the size of MiniImagenet.
    \item Plotting the meta-generalization gap (with a synthetic classification task) and demonstrate it decreases as the number of tasks increases would be interesting (note however we already have the limiting case when the number of tasks is unbounded and the meta-generalization gap is zero).
    \item An interesting experiment would be to train a deep neural network with the episodic training (but without the MAML inner loop) but have an unbounded number of tasks and see if the test error keeps increases (or stays at change, as observed when this is done with MiniImagenet \ref{maml0}).
    \item An interesting hypothesis to investigate is if meta-learning algorithms get representation that are optimal for their respective meta-learner (or adaptation rule).
    If this is true, it means methods like \cite{Tian2020} can be improved by making the entire pipeline differential and learning it end-to-end \cite{dimo}.
    \item Test meta-learning algorithms in domains where higher level cognition is required and thus compositionality is essential, e.g. program synthesis \cite{Chen2020} and theorem proving \cite{sketch, trail}.
    % \item Deliberately design an AGI and AI safety measure as proposed in \cite{foundationsmetalearning} for few-shot learning and emperically test them.
    % \item Propose a robust and widely accepted general intelligence metric that is applicable for many environments and tasks - in particular for few-shot learning.
    % We believe deliberate efforts for general intelligence are important.
\end{enumerate}

\subsubsection{Proposal on Synthetic classification task that possibly require meta-learning}

Synthetic tasks that use classification instead of regression are not hard to define.
Two possible alternatives are: 
1) a mixture of Gaussian but the standard deviation controls the radius of limit of how far the classes can be from each other
2) another option is the similar as with a mixture of Gaussian but have 
the (vector) samples be weights of a Neural network (so that the goal is to identify from which Neural Network data is coming from)

\section{Broader Impact}

% \sk{I suggest removing this section -- its too far disconnected from the evidence and main results actually presented, and it will reduce reader confidence. This can work for talks, or position pieces, but a technical paper needs a stronger link between topics your cover than what you have here. GI is good, but meta-learning is only a small piece of it. Metrics are good, but your work is really about meta-learning, not a treatise on metrics. }
\subsection{Quantifying general intelligence through meta-learning}
There are valuable efforts that try to make benchmarks which require higher level cognition, e.g. \cite{Vedantam}.
An example of work that tries to quantify AGI and proposes a benchmark is \cite{Chollet2019}.
We believe the second approach is likely to have more impact in the long run because it also deliberately quantifies general intelligence.
We believe that suggesting benchmarks without clearly specifying the long term goal or measuring the metric we are trying to optimize is a suboptimal approach.
However, we do believe grounding benchmarks on tasks that humans are able to perform is a good idea but suggest augmenting these proposals with metrics and explicit discussions of general intelligence.

Another approach we believe has high potential is program synthesis \cite{Chen2020} and theorem proving \cite{sketch, trail} because humans create higher abstractions that are composed and re-use, thus suggesting to meta-learning might be taking place.
We believe that higher level cognition tasks are a challenging to assess meta-learning algorithms.

\subsection{Quantifying AI safety}
We also believe quantifying and tracking metrics for AI safety as early as possible is crucial.
Few-shot learning is likely one of the simplest - and arguably the atomic buildings blocks for general intelligence. 
We believe AI safety could be enriched if the research community deliberately tracks, discusses and report it in all its research - especially in meta-learning research.
For a brief discussion, see \cite{foundationsmetalearning}.

\newpage

\subsection{Summary of Broader Impact}
% \sk{same here, the link is too tenuous, and claims are too general IMHO}
We hope that this discussion inspires the AI community - but especially the meta-learning research community - to always report their progress deliberately using, what Miranda et al. \cite{foundationsmetalearning} call the "the big three": 
\begin{enumerate}
    \item the score for absolute performance (to ensure usefulness)
    \item the score for general skill acquisition (to ensure flexibility and general intelligence)
    \item the AI safety score (to ensure positive outcome)
\end{enumerate}

\end{document}